\documentclass[10pt,twocolumn,letterpaper]{article}

\usepackage[pagenumbers]{iccv} 
\usepackage{kotex}
\usepackage{subcaption}
\usepackage{algorithm}
\usepackage{algorithmicx}
\usepackage{algpseudocode}
\usepackage{amsmath}
\usepackage{multirow}
\usepackage{graphicx}
\usepackage[normalem]{ulem}
\usepackage{xcolor}
\usepackage{amssymb}
\usepackage{hhline}
\usepackage{colortbl}
\usepackage{xcolor}
\newcommand{\circled}[1]{\textcircled{\scriptsize #1}}

\definecolor{mycolor}{HTML}{DBE2EF} 
\definecolor{mycolor2}{HTML}{EAEAEA} 

\usepackage[sectionbib]{chapterbib} 

\definecolor{iccvblue}{rgb}{0.21,0.49,0.74}
\usepackage[pagebackref,breaklinks,colorlinks,allcolors=iccvblue]{hyperref}

\def\paperID{9190} 
\def\confName{ICCV}
\def\confYear{2025}

\title{GranQ: Efficient Channel-wise Quantization via Vectorized Pre-Scaling for Zero-Shot QAT}

\author{Inpyo Hong, Youngwan Jo, Hyojeong Lee, Sunghyun Ahn, Kijung Lee, Sanghyun Park\\
Yonsei University\\
{\tt\small \{hip9863,jyy1551, hyojoy, skd, rlwjd4177, sanghyun\}@yonsei.ac.kr}}

\begin{document}

\maketitle

\begin{abstract}
Zero-shot quantization (ZSQ) enables neural network compression without original training data, making it a promising solution for restricted data access scenarios. To compensate for the lack of data, recent ZSQ methods typically rely on synthetic inputs generated from the full-precision model. However, these synthetic inputs often lead to activation distortion, especially under low-bit settings. To mitigate this, existing methods typically employ per-channel scaling, but they still struggle due to the severe computational overhead during the accumulation process. To overcome this critical bottleneck, we propose \textit{GranQ}, a novel activation quantization framework that introduces an efficient pre-scaling strategy. Unlike conventional channel-wise methods that repeatedly perform scaling operations during accumulation, \textit{GranQ} applies scaling factors in a pre-scaling step through fully vectorized computation, eliminating runtime scaling overhead. This design enables \textit{GranQ} to maintain fine-grained quantization accuracy while significantly reducing computational burden, particularly in low-bit quantization settings. Extensive experiments under quantization-aware training (QAT) settings demonstrate that \textit{GranQ} consistently outperforms state-of-the-art ZSQ methods across CIFAR and ImageNet. In particular, our method achieves up to 5.45\% higher accuracy in the 3-bit setting on CIFAR-100 and even surpasses the full-precision baseline on CIFAR-10.
\end{abstract}

\begin{table}[h]
\resizebox{0.46\textwidth}{!}{%
\centering
\begin{tabular}{ccc} 
\hline
\textbf{\begin{tabular}[c]{@{}c@{}}Data Generation\\ (PTQ, \textcolor{violet}{QAT})\end{tabular}} &
  \textbf{\begin{tabular}[c]{@{}c@{}}Calibration\\ (PTQ)\end{tabular}} &
  \textbf{\begin{tabular}[c]{@{}c@{}}Fine-tuning\\ (QAT)\end{tabular}} \\ \hline
\begin{tabular}[c]{@{}c@{}}ZeroQ \textit{(CVPR 20)}\\ GDFQ \textit{(ECCV 20)}\\ DSG \textit{(CVPR 21)}\\ MixMix \textit{(CVPR 21)}\\ Qimera \textit{(NeurIPS 21)}\\ IntraQ \textit{(CVPR 22)}\\ GENIE \textit{(CVPR 23)}\\ \textcolor{violet}{AdaDFQ \textit{(CVPR 23)}}\\ \textcolor{violet}{Casual-DFQ \textit{(ICCV 23)}}\\ \textcolor{violet}{TexQ \textit{(NeurIPS 23)}}\\ \textcolor{violet}{RIS \textit{(AAAI 24)}}\\ GenQ \textit{(ECCV 24)}\end{tabular} &
\begin{tabular}[c]{@{}c@{}}SQuant \textit{(ICLR 22)}\\ UDFC \textit{(ICCV 23)}\end{tabular} &
\begin{tabular}[c]{@{}c@{}}AIT \textit{(CVPR 22)}\\ PLF \textit{(CVPR 24)} \\  AKT \textit{(SAC 25)}\\ SynQ \textit{(ICLR 25)}\\ \textbf{GranQ \textit{(Ours)}}\end{tabular} \\ \hline 
\end{tabular}%
}
\caption{Categorizing ZSQ algorithms. PTQ denotes post-training quantization, and QAT denotes quantization-aware training. Data generation methods typically support both PTQ and QAT, whereas \textcolor{violet}{violet-marked methods} utilize fine-tuning for data generation and \textcolor{violet}{support only QAT.}}
\label{tab1}
\end{table}

\begin{figure*}[t]
    \centering
    
    \begin{subfigure}[b]{0.47\textwidth}
        \centering
        \includegraphics[width=\textwidth]{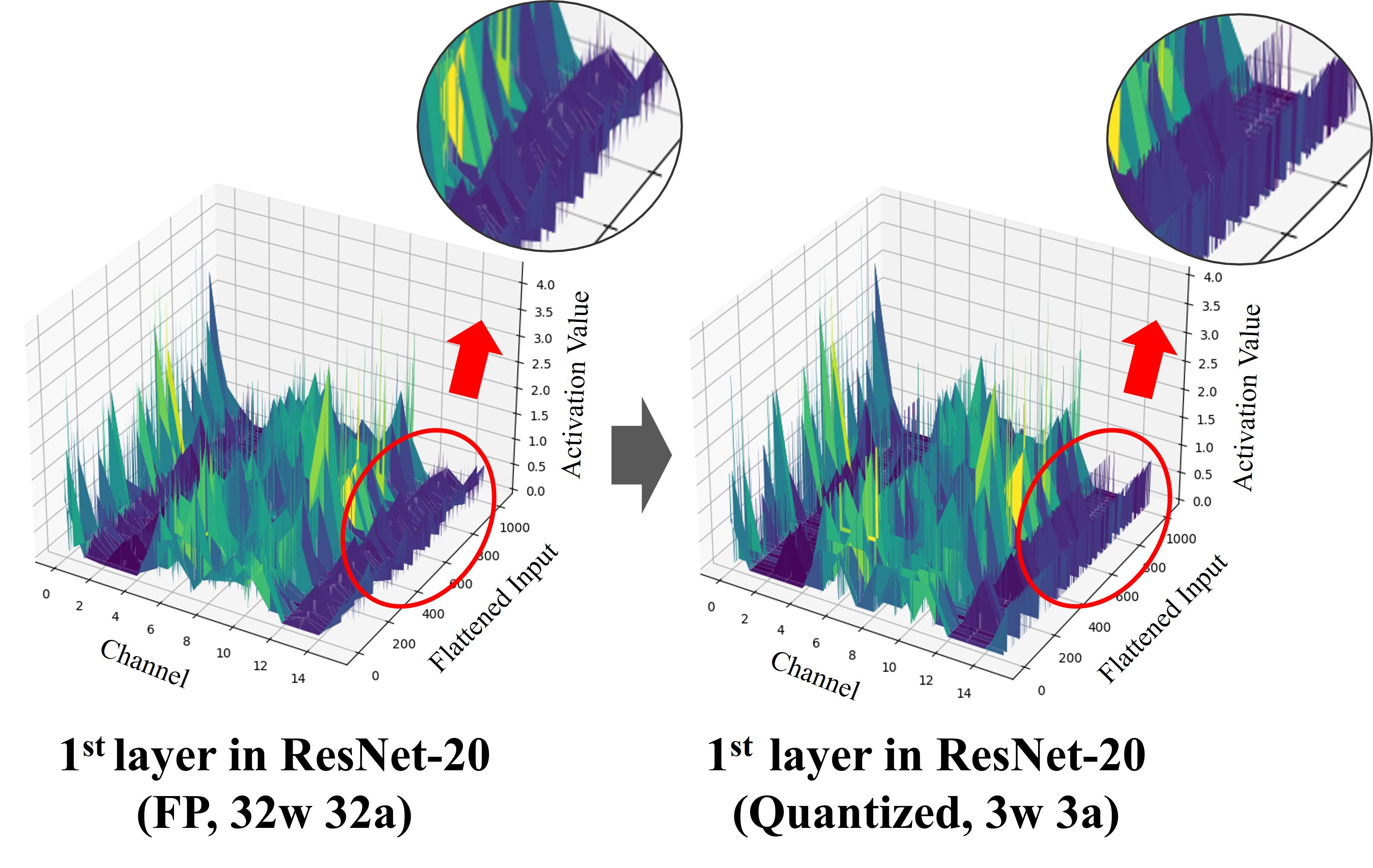}
        \caption{Layer-wise Quantization}
        \label{fig1-1}
    \end{subfigure}
    \hfill
    \begin{subfigure}[b]{0.47\textwidth}
        \centering
        \includegraphics[width=\textwidth]{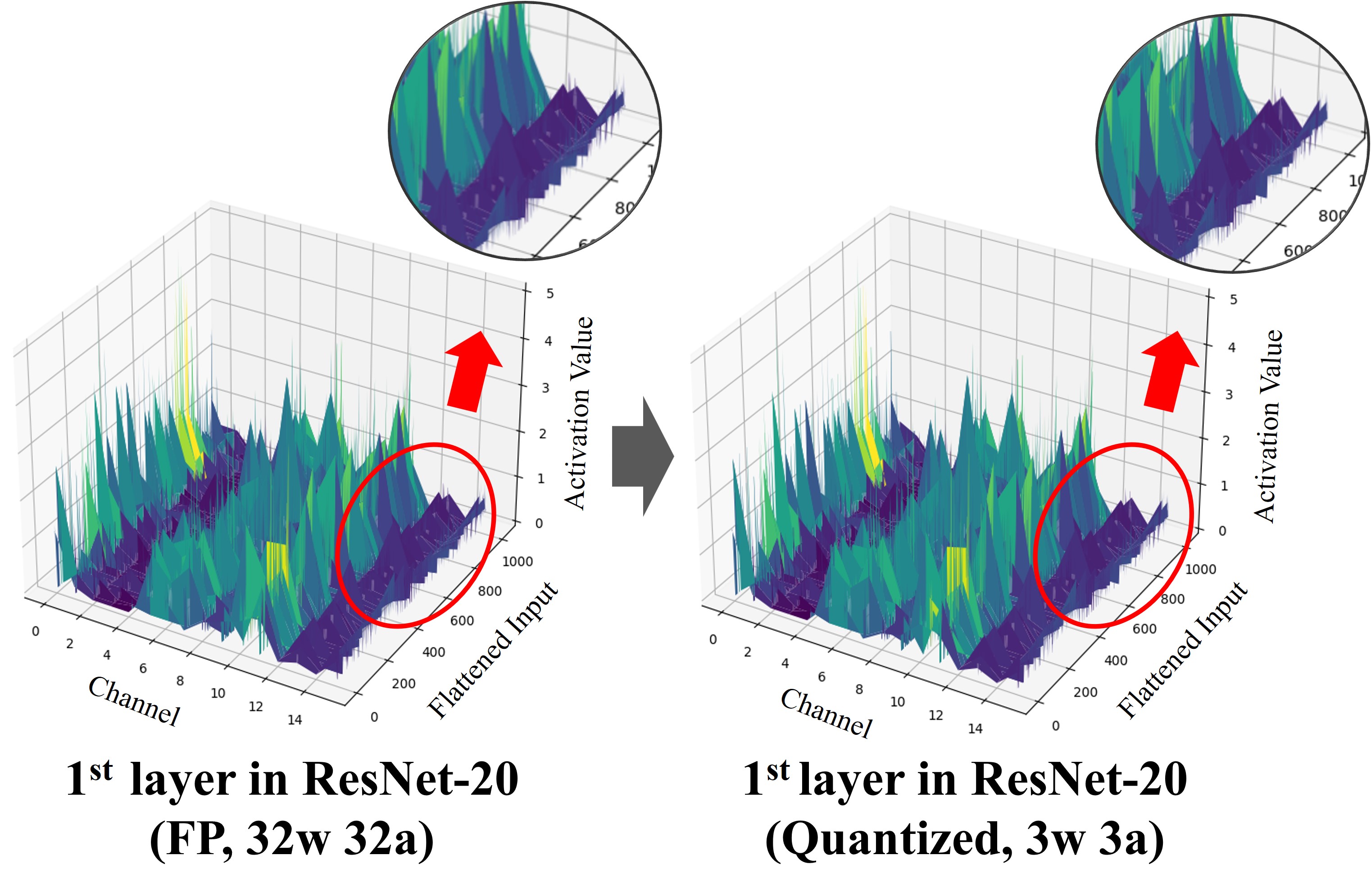}
        \caption{\textit{GranQ} (Channel-wise, Ours)}
        \label{fig1-2}
    \end{subfigure}
    \caption{Comparison between (a) layer-wise quantization and (b) \textit{GranQ} on the CIFAR-10. Each subfigure visualizes the 32-bit FP (left) and 3-bit quantized (right) activations of the first ResNet-20 layer. \textit{GranQ} better preserves the original activation with minimal distortion.}
    \label{fig1}
    \vspace{-0.2em}
\end{figure*}

\section{Introduction}
\label{sec:intro}

Neural network compression has been extensively studied for the practical deployment of large-scale deep learning (DL) models. In particular, reducing model size while minimizing performance degradation is crucial for utilizing DL on edge devices (e.g., mobile phones, embedded systems, and drones). Major approaches to model compression include quantization~\cite{3,4,5}, pruning~\cite{6,7,8,9}, knowledge distillation~\cite{10,11,12}, and neural architecture search~\cite{13,14}, as surveyed in~\cite{1,2}. Among these, quantization has emerged as the most actively studied technique. It serves as an effective compression method by reducing unnecessary representational ranges in the model. However, it often requires fine-tuning or calibration to match full-precision (FP) model performance \cite{5}. To address this, zero-shot quantization (ZSQ), also known as data-free quantization, has been proposed to compress models without the original training data.

Since the introduction of ZeroQ \cite{17}, studies on ZSQ have advanced in two main directions. The first direction focuses on data generation, where synthetic data are created from the FP model. The second direction focuses on effectively applying the activation distributions of the synthetic data to the quantized (Q) model. This second direction is further divided into post-training quantization (PTQ), which calibrates the activation distributions, and quantization-aware training (QAT), which fine-tunes the Q model directly. We categorize existing ZSQ methods based on these directions, as summarized in Table \ref{tab1}.

First, data generation studies focus on generating high-quality data to effectively train the Q model \cite{16, 22, 26}. Meanwhile, calibration (PTQ) studies aim to minimize the quantization error by calibrating the Q model without additional training \cite{31,32}. Finally, fine-tuning (QAT) studies focus on transferring key information from the FP model to the Q model through knowledge distillation \cite{27,28,29,30}. 

However, despite extensive studies, severe performance degradation in low-bit quantization remains unresolved. To address this issue, we performed an in-depth analysis of the ZSQ process, focusing on why low-bitwidth settings still suffer from performance loss even after QAT fine-tuning. Our findings reveal that quantization errors mainly stem from the loss of activation values instead of data quality or training methods. Notably, we found that layer-wise (per-tensor) quantization is no longer suitable for preserving activations in ZSQ, as it leads to coarse and inaccurate representations.

Based on this analysis, we introduce \textit{GranQ}, a novel ZSQ method that achieves efficient per-channel quantization via vectorized pre-scaling of input-dependent activations. This dynamic adjustment minimizes activation loss and preserves the original activation values by reducing quantization errors, as shown in Figure \ref{fig1}. The proposed method effectively handles activation loss in low-bit quantization and achieves state-of-the-art (SOTA) performance in QAT settings on the CIFAR and ImageNet datasets. Furthermore, we apply vectorization to the pre-scaling step, which is typically omitted in conventional channel-wise quantization, thereby reducing latency and enabling fine-grained activation quantization. Our contributions can be summarized as follows:



\begin{itemize}
\item \textbf{We identify critical limitations of layer-wise activation quantization in low-bit ZSQ.} Our findings reveal that conventional activation quantization methods relying on layer-wise granularity suffer from significant activation loss. This limitation becomes more severe in ZSQ settings with synthetic data.

\item \textbf{We propose \textit{GranQ}, a novel method that supports granular quantization and maintains computational efficiency.}
Although per-channel activation quantization is known to improve precision, it has not been widely adopted due to its high computational cost. We address this by introducing vectorized pre-scaling, which integrates per-channel scaling into the quantization step, allowing accumulation to proceed efficiently without runtime scaling overhead. To the best of our knowledge, our approach is the first to address the ZSQ problem.

\item \textbf{We achieve SOTA performance over existing ZSQ methods through extensive evaluation.} Specifically, on the CIFAR-100 dataset, in the 3-bit quantization setting, \textit{GranQ} achieves an accuracy of 62.73\%, improving by 5.45\% over the latest method on ResNet-20. Furthermore, on the CIFAR-10 dataset in the 5-bit quantization setting, \textit{GranQ} achieves an accuracy of 94.06\%, slightly exceeding the FP model performance by 0.17\%.
\end{itemize}

\section{Related Work}
\label{sec:Related Work}

\subsection{Quantization}

Quantization reduces the representational range of deep neural networks (DNNs) and minimizes memory usage. It is typically categorized into post-training quantization (PTQ) and quantization-aware training (QAT) \cite{5, 34}. PTQ applies quantization after training without further updates, which makes it efficient and easy to implement. However, it is sensitive to scaling errors and often relies on calibration to improve accuracy. In contrast, QAT incorporates quantization during training and optimizes quantized activations using the straight-through estimator \cite{5, 35}. Both methods require data. PTQ uses it for calibration, while QAT uses it for fine-tuning.. In practice, access to training data is often limited. To address this, zero-shot quantization (ZSQ) has been proposed to perform quantization without using original data \cite{16}.

\subsection{Zero-shot Quantization}
As summarized in Table \ref{tab1}, ZSQ was initially introduced by ZeroQ \cite{17}, leading to the development of various ZSQ algorithms. These studies have explored diverse approaches to improve quantization performance under data-free settings. Recent studies have focused on generating high-quality synthetic data \cite{23, 26} and developing more effective methods to utilize them \cite{27, 29, 30}. However, current ZSQ methods heavily rely on augmented synthetic inputs. This causes large variation in activation scales across channels. In practice, such input-dependent channel-wise quantization is rarely adopted due to its high computational cost. Therefore, recent studies have applied channel-wise activation quantization either in a limited conditions within QAT settings \cite{47}, or exclusively in PTQ scenarios with access to a small amount of data \cite{48,49}. This highlights the importance of efficient channel-wise activation quantization in QAT.

\section{Preliminaries and Problem Definition}
\label{sec:preliminaries}
In this section, we provide a new definition of the existing ZSQ problem and introduce the preliminaries. 



\subsection{Activation Quantization}
Activation quantization reduces the precision of intermediate activation values by converting them into low-bitwidth integers \cite{36}. It is commonly used with weight quantization to compress models, and the linear quantization scheme is most widely adopted. This process typically consists of quantization, scaling parameter computation, and dequantization \cite{3, 5, 37}.

The quantization operator $\mathcal{Q}$ maps a floating-point value $x$ into an integer $x_q$ using a scaling factor $s$ and zero-point $z$:
\begin{equation}
x_q = \mathcal{Q}(x, s, z) = \left\lfloor \frac{x}{s} + z \right\rceil
\label{eq1}
\end{equation}
Here, $s$ normalizes the activation range, and $z$ shifts the scaled value to account for asymmetric quantization. These parameters are computed as:
\begin{equation}
s = \frac{x_{\max} - x_{\min}}{2^b - 1}, \quad z = \left\lfloor \frac{-x_{\min}}{s} \right\rceil
\label{eq2}
\end{equation}
where $x_{\min}$ and $x_{\max}$ denote the range of activation values, and $b$ is the bit-width. A large range yields a large $s$, increasing quantization error, a smaller range allows finer resolution.
As shown below, in quantized models, per-channel scaling is embedded directly into the accumulation path, requiring each channel to be individually scaled during computation:
\begin{equation}
y_l = \sum_{c=1}^{C} w_c \cdot s_c \cdot (x_{q,c} - z_c)
\label{eq3}
\end{equation}

However, applying the per-channel scaling factor $s_c$ inside the accumulation loop introduces computational overhead, making efficient integer-domain execution difficult. This scaling bottleneck hinders parallelism in core operations such as convolution and matrix multiplication, reducing the efficiency of quantized models \cite{50}.

\subsection{Problem Definition}

Existing activation quantization methods are designed based on layer-wise quantization to minimize computational cost. However, a fixed scaling factor struggles to handle varying activations. This issue becomes more severe under low-bitwidth settings in ZSQ environments. The following outlines the primary challenges faced in ZSQ under low-bitwidth settings.\\
\textbf{\textit{(Problem 1)}} \textbf{Coarse Activation Quantization by Single-Range Scaling}\\
Conventional activation quantization uses a single scaling range per layer for efficiency, which was acceptable at higher bitwidths or when real data is available. However, this design becomes problematic in ZSQ, where range adjustment relies on synthetic data. These synthetic distributions are often biased or irregular, making precise quantization more difficult.\\
\textbf{\textit{(Problem 2)}} \textbf{Accumulation Bottleneck in Channel-wise Quantization}\\
The main challenge in channel-wise quantization lies not in quantizing activations but in the repeated per-channel scaling during accumulation. This disrupts vectorized execution and significantly increases computational overhead.


\section{Observation and Methodology}
\label{sec:methodology}
We observe the causes of activation distortion during quantization and analyze how to mitigate them. Based on this, we introduce our method, \textit{GranQ}.



\subsection{Observation}

\begin{table}[b]
\centering
\renewcommand{\arraystretch}{1.3}
\resizebox{0.5\textwidth}{!}{%
\begin{tabular}{ccc}
\hline
\textbf{Activation Quantization} & \textbf{Layer-wise} & \textbf{Channel-wise (\textit{GranQ})} \\ \hline
\begin{tabular}[c]{@{}c@{}} 
{\small $Avg.\frac{X \cdot Q}{\|X\|_2 \|Q\|_2}(\uparrow)$} \\ 
{\small $Avg.\frac{\| X - Q \|_2}{\| X \|_2} (\downarrow)$} 
\end{tabular} 
& 
\begin{tabular}[c]{@{}c@{}} 
{\small 0.5111} \\ {\small 0.3129} 
\end{tabular} 
& 
\begin{tabular}[c]{@{}c@{}} 
{\small  \cellcolor{mycolor}\textbf{0.6835}} \\ {\small  \cellcolor{mycolor}\textbf{0.1063}} 
\end{tabular} 
\\ \hline 
\end{tabular}%
}
\caption{Average cosine similarity and relative error for 3-bit activation quantization on ResNet-20 (CIFAR-100). \textit{GranQ} shows lower activation distortion than layer-wise, with higher similarity and lower error.}
\label{tab2}

\end{table}

\begin{figure*}[t]
    \centering
    \includegraphics[width=0.98\textwidth]{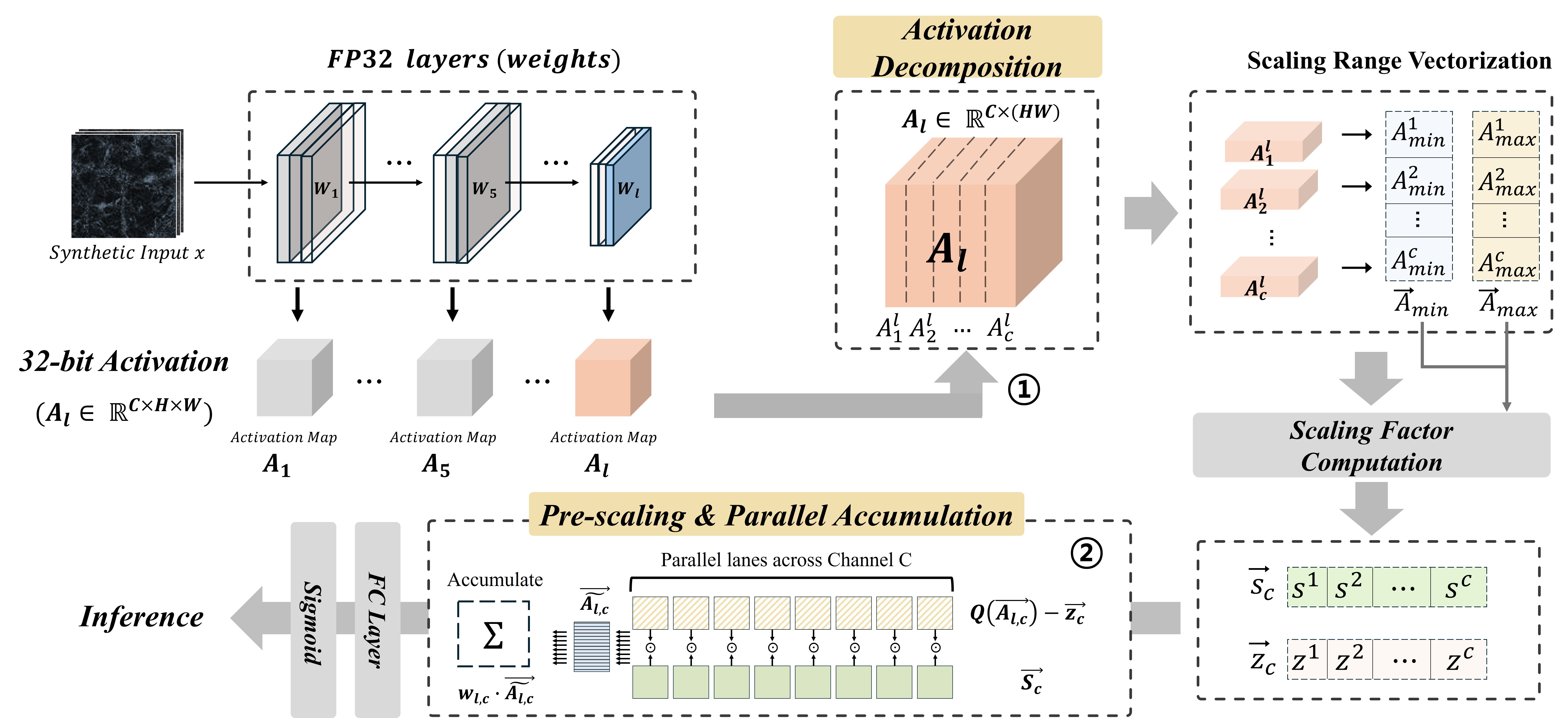}
    \caption{Overview of the \textit{GranQ} algorithm. \circled{1} Each activation map $A_l$ is decomposed into channel-wise vectors, which are used to compute scaling factors ($\vec{s}_c$) and zero-points ($\vec{z}_c$) in a vectorized form. \circled{2} The calculated scaling factor is applied in advance (pre-scaled) to the quantized activations through parallel lanes, enabling efficient parallel accumulation.
}

    \label{fig3}
\end{figure*}
Layer-wise quantization effectively captures the activation distribution of each layer. However, the effectiveness of this layer-wise approach is compromised by high inter-channel variance in activations. Particularly in ZSQ, as shown in Figure~\ref{fig1-1}, the activation magnitudes exhibit a large variance along the channel axis. This implies that limited representation bits are forced to capture a wide range of activation values, making fine-grained quantization difficult. This phenomenon is not isolated to a single layer but appears consistently throughout the model, as summarized by our analysis in Table \ref{tab2}. It presents the cosine similarity $\frac{X \cdot Q}{|X|_2 |Q|_2}$, along with the relative error $\frac{| X - Q |_2}{| X |_2}$. From the results, we observe that layer-wise quantization fails to preserve activation information effectively. \textbf{This reveals a critical limitation of using a single scaling range for activation quantization, particularly in ZSQ.} In contrast, \textit{GranQ}, which applies scaling per channel, achieves 1.34$\times$ higher similarity and 2.94$\times$ lower relative error than layer-wise quantization, demonstrating its effectiveness in preserving activation information.

\begin{algorithm}[b]
\footnotesize
\caption{\textit{GranQ}: Pre-scaling for Parallel Accumulation}
\label{alg1}
\begin{algorithmic}[1]
\Require Activation tensors $\{A_l\}_{l=1}^L$, where $A_l \in \mathbb{R}^{C \times H \times W}$, quantization bit $b$, weight tensors $\{w_{l,c}\}_{l=1,c=1}^{L,C}$
\Ensure Output vectors $\{y_l\}_{l=1}^L$
\For{each layer $l$ in $L$}
    \State $A_l \gets \text{reshape}(A_l) \in \mathbb{R}^{C \times (HW)}$
    \State $\vec{A}_{\min}, \vec{A}_{\max} \gets \min_{h,w}(A_l), \max_{h,w}(A_l)$
    \State $\vec{s}_l \gets (\vec{A}_{\max} - \vec{A}_{\min}) / (2^b - 1)$
    \State $\vec{z}_l \gets \left\lfloor -\vec{A}_{\min} / \vec{s}_l \right\rceil$
    \State $\vec{A}_{q,l} \gets \left\lfloor A_l / \vec{s}_l + \vec{z}_l \right\rceil$ \Comment{\textit{quantize (int mapping)}}
    \State $\vec{\tilde{A}}_{l,c} \gets \vec{s}_l \odot (\vec{A}_{q,l} - \vec{z}_l)$ \Comment{\textit{dequantize \& pre-scale}}
    \State \textbf{Compute output:}
    \[
    y_l = \sum_{c=1}^{C} w_{l,c} \cdot \vec{\tilde{A}}_{l,c}
    \]
\EndFor
\State \Return $\{y_l\}_{l=1}^L$
\end{algorithmic}
\end{algorithm}

\subsection{Methodology}

Based on our analysis, we propose \textit{GranQ}, a fine-grained ZSQ method designed to reduce computational overhead, as illustrated in Figure~\ref{fig3}. \textit{GranQ} applies channel-wise scaling through \textit{activation decomposition}, and achieves fast parallel processing via \textit{pre-scaling for parallel accumulation}, which enables efficient computation of both scaling and quantization.


\subsubsection{Activation Decomposition}

We propose \textit{activation decomposition}, which reshapes each activation map from a three-dimensional tensor of shape $(C \times H \times W)$ into a two-dimensional matrix of shape $(C \times HW)$. This transformation enables decomposition of the activation map along the channel axis, allowing each channel to be processed independently. It also facilitates vectorized computation of per-channel statistics (e.g., min and max) required for scaling factor calculation during quantization. We define the \textit{activation decomposition} as shown in Equation~\ref{eq4}.


\begin{equation}
\begin{gathered}
    A_l = \left[ A_l(1,:,:),\, A_l(2,:,:),\, \dots,\, A_l(C,:,:) \right] \in \mathbb{R}^{C \times H \times W} \\
    \vec{A}_{\min} = \left[ \min_{(h,w)} A_l(c,h,w) \right]_{c=1}^C \in \mathbb{R}^C \\
    \vec{A}_{\max} = \left[ \max_{(h,w)} A_l(c,h,w) \right]_{c=1}^C \in \mathbb{R}^C
\end{gathered}
\label{eq4}
\end{equation}

Here, the vectors $\vec{A}_{\min}$ and $\vec{A}_{\max}$ denote the channel-wise minimum and maximum values computed from the activation input $A_l$ of each layer, where $A_l \in \mathbb{R}^{C \times H \times W}$. Unlike traditional layer-wise quantization that applies a single scalar across all channels, this representation allows for independent normalization of each channel. 





\begin{table*}[t]

\centering
\renewcommand{\arraystretch}{1.4}
\resizebox{1.0\textwidth}{!}{%
\begin{tabular}{c|c|c|cccccccc|>{\columncolor{mycolor}}c} 

\hline
\textbf{Dataset} & \textbf{\begin{tabular}[c]{@{}c@{}}Model\\ (FP 32)\end{tabular}} & \textbf{Bits} 
& \textbf{\begin{tabular}[c]{@{}c@{}}GDFQ\\ \textit{(ECCV 20)}\end{tabular}} 
& \textbf{\begin{tabular}[c]{@{}c@{}}ARC+AIT\\ \textit{(CVPR 22)}\end{tabular}} 
& \textbf{\begin{tabular}[c]{@{}c@{}}AdaDFQ\\ \textit{(CVPR 23)}\end{tabular}} 
& \textbf{\begin{tabular}[c]{@{}c@{}}TexQ\\ \textit{(NeurIPS 24)}\end{tabular}} 
& \textbf{\begin{tabular}[c]{@{}c@{}}AIT+RIS\\ \textit{(AAAI 24)}\end{tabular}} 
& \textbf{\begin{tabular}[c]{@{}c@{}}GenQ\\ \textit{(ECCV 24)}\end{tabular}} 
& \textbf{\begin{tabular}[c]{@{}c@{}}AKT\\ \textit{(SAC 25)}\end{tabular}} 
& \textbf{\begin{tabular}[c]{@{}c@{}}SynQ\\ \textit{(ICLR 25)}\end{tabular}} 
& \textbf{\begin{tabular}[c]{@{}c@{}}\cellcolor[HTML]{ffffff}\textit{GranQ}\\ \cellcolor[HTML]{ffffff}\textit{(Ours)}\end{tabular}} \\ \hline
\multirow{3}{*}{Cifar-10} & \multirow{3}{*}{\begin{tabular}[c]{@{}c@{}}ResNet-20\\ (93.89)\end{tabular}} & 3$w$3$a$ 
& 75.11 & - & 84.89 & 86.47 & - & - & 86.76 & \underline{88.11} & \textbf{91.37} \\
 &  & 4$w$4$a$ & 90.25 & 90.49 & 92.31 & 92.68 & 92.59 & - & 92.64 & \underline{92.76} & \textbf{93.52} \\
 &  & 5$w$5$a$ & 93.38 & 92.98 & 93.81 & - & 93.59 & - & \underline{93.83} & - & \textbf{94.06} \\ \hline
\multirow{3}{*}{Cifar-100} & \multirow{3}{*}{\begin{tabular}[c]{@{}c@{}}ResNet-20\\ (70.33)\end{tabular}} & 3$w$3$a$ 
& 47.61 & - & 52.74 & 55.87 & - & - & 54.68 & \underline{57.28} & \textbf{62.73} \\
 &  & 4$w$4$a$ & 63.39 & 61.05 & 66.81 & 67.18 & 65.99 & - & 66.94 & \underline{67.34} & \textbf{68.79} \\
 &  & 5$w$5$a$ & 66.12 & 68.40 & \underline{69.93} & - & 69.55 & - & 69.75 & - & \textbf{70.05} \\ \hline
\multirow{9}{*}{ImageNet} & \multirow{3}{*}{\begin{tabular}[c]{@{}c@{}}ResNet-18\\ (71.47)\end{tabular}} & 3$w$3$a$ 
& 20.23 & - & 38.10 & 50.28 & - & \textbf{68.18} & 49.88$^\dagger$ & 52.02 & \underline{64.41} \\
 &  & 4$w$4$a$ & 60.60 & 65.73 & 66.53 & 67.73 & 67.55 & \underline{70.03} & 65.89$^\dagger$ & 67.90 & \textbf{70.39} \\
 &  & 5$w$5$a$ & 68.49 & 70.28 & 70.29 & - & \underline{70.59} & - & 69.40$^\dagger$ & - & \textbf{71.31}  \\  \hhline{~-----------}
 & \multirow{3}{*}{\begin{tabular}[c]{@{}c@{}}MobileNetV2\\ (73.03)\end{tabular}} & 3$w$3$a$ 
& 1.46 & - & 28.99 & 32.80 & - & \underline{59.15} & 30.56$^\dagger$ & 34.21 & \textbf{62.42} \\
 &  & 4$w$4$a$ & 59.43 & 66.47 & 65.41 & 67.07 & - & \underline{69.65} & 64.85$^\dagger$ & 67.27 & \textbf{70.62} \\
 &  & 5$w$5$a$ & 68.11 & \underline{71.96} & 71.61 & - & - & - & 71.71$^\dagger$ & - & \textbf{72.49} \\ \hhline{~-----------}
 & \multirow{3}{*}{\begin{tabular}[c]{@{}c@{}}ResNet-50\\ (77.73)\end{tabular}} & 3$w$3$a$ 
& 0.31 & - & 17.63 & 25.27 & - & \textbf{73.99} & 24.50$^\dagger$ & 26.89 & \underline{70.76} \\
 &  & 4$w$4$a$ & 54.16 & 68.27 & 68.38 & 70.72 & 71.54 & \underline{76.10} & 68.75$^\dagger$ & 71.05 & \textbf{76.63} \\
 &  & 5$w$5$a$ & 71.63 & 76.00 & 76.03 & - & \underline{76.36} & - & 75.90$^\dagger$ & - & \textbf{77.58} \\ \hline 
\end{tabular} }
\caption{Accuracy Evaluation of QAT Methods for ZSQ. $w$ and $a$ represent weight and activation, respectively. \textbf{Bold} values indicate the best accuracy, and \underline{underlined} values denote the second-best accuracy. $\dagger$ indicates our re-implementation.}
\label{tab3}
\vspace{-1em}
\end{table*}

\subsubsection{Pre-scaling for Parallel Accumulation}

\begin{equation}
    \vec{s} = \frac{\vec{A}_{\max} - \vec{A}_{\min}}{2^b - 1}, \quad
    \vec{z} =  \left\lfloor - \frac{\vec{A}_{\min}}{\vec{s}} \right\rceil
    \label{eq5}
\end{equation}

\begin{equation}
    y_l = \sum_{c=1}^{C} w_{l,c} \cdot \vec{\tilde{A_{l,c}}}, \quad 
    \text{where } \vec{\tilde{A_{l,c}}} = \vec{s}_c \odot \left( \left\lfloor \frac{\vec{A}_{l,c}}{\vec{s}_c} + \vec{z}_c \right\rceil - \vec{z}_c \right)
    \label{eq6}
\end{equation}

To fully leverage the capabilities of modern vectorized hardware for quantization, we introduce the \textit{pre-scaling for parallel accumulation} stage, a novel algorithmic framework. Within this  stage, the scaling factors $\vec{s}_c$ and zero-points $\vec{z}_c$ are computed in a vectorized form across channels to dynamically adapt to the diverse activation distributions induced by synthetic inputs. This contrasts with conventional per-channel quantization, which applies fixed scaling parameters per channel regardless of input variation.

As shown in Equation~\ref{eq6}, each activation $\vec{A}_{l,c}$ is first quantized using its corresponding vectorized scaling and zero-point, and then dequantized to reconstruct the floating-point value. The resulting $\vec{\tilde{A_{l,c}}}$ represents a pre-scaled activation that is already adjusted for accumulation. During dequantization, a hadamard product ($\odot$) is used to perform an element-wise multiplication between the vectorized scaling factor $s_c$ and the quantized activation. Crucially, the element-wise nature of this operation means that the computation for each channel is independent of the others. This property allows the pre-scaling of all channels to be executed simultaneously across dedicated hardware lanes, as illustrated in part \circled{2} of Figure \ref{fig3}. By applying vectorized scaling, quantization, and dequantization uniformly across channels, this process enables efficient integer-domain computation and parallel accumulation with weights $w_{l,c}$. Our core innovation is an algorithmic re-formulation of the quantization process. We mathematically restructure the operation to isolate the heavy accumulation step as a pure integer-domain task. This design unlocks the full potential of parallel hardware, achieving a level of efficiency unattainable by conventional methods.

\section{Experiments}
\label{sec:experiments}

In this section, we thoroughly evaluate the effectiveness of GranQ. Experiments are conducted on diverse benchmark datasets, with the performance compared with those of existing ZSQ methods.


\subsection{Experimental Setup and Details}

The experiments were conducted using widely adopted ZSQ evaluation datasets, including CIFAR-10, CIFAR-100 \cite{40}, and ImageNet (ILSVRC 2012) \cite{41} validation datasets. For the CIFAR datasets, ResNet-20 \cite{42} was used as the quantization model, whereas ResNet-18 \cite{42}, ResNet-50 \cite{42}, and MobileNetV2 \cite{43} were employed for ImageNet.
All experiments were conducted using the SGD optimizer \cite{44} with a momentum of 0.9 and weight decay of 1e-4. The CIFAR-10 and CIFAR-100 experiments were each conducted for 200 epochs, with batch sizes of 16 and 200, respectively. For ImageNet, we trained for 400 epochs with a batch size of 16. The initial learning rate was set to 1e-4 for CIFAR-10 and CIFAR-100, and 1e-5 for ImageNet, with multi-step learning rate decay applied. The decay steps were set to 100, 200, and 300 epochs for CIFAR, and at 350 and 400 epochs for ImageNet, with a decay rate of 0.1.
We compared our method with existing ZSQ methods \cite{16, 27, 22, 24, 25, 28, 26, 29, 30}. For data generation, we followed the AdaDFQ \cite{22} approach based on ACGAN \cite{45}. Layer-wise quantization was applied to all layers containing activation functions, while channel-wise quantization was performed per channel at the batch level. For the quantization scheme, we applied channel-wise quantization for all weights.

\begin{table*}[t]
\centering
\resizebox{0.9\textwidth}{!}{%
\begin{tabular}{c|cc|cc|cc|cc} 
\hline
\multicolumn{1}{c|}{\textbf{CIFAR-100}} & \multicolumn{8}{c}{\textbf{ResNet-20 (70.33\%)}} \\ \hline
\multirow{2}{*}{\textbf{Method}} & \multicolumn{2}{c|}{\textbf{GDFQ}} & \multicolumn{2}{c|}{\textbf{Qimera+AIT}} & \multicolumn{2}{c|}{\textbf{AdaDFQ}} & \multicolumn{2}{c}{\textbf{AdaDFQ+AKT}} \\ \cline{2-9}
 & Baseline & +\textit{GranQ} & Baseline & +\textit{GranQ} & Baseline & +\textit{GranQ} & Baseline & +\textit{GranQ} \\ \hline
3$w$3$a$ & 47.61    & \cellcolor{mycolor}59.04$^{+11.43}$  & 45.70$^\dagger$    & \cellcolor{mycolor}60.42$^{+14.72}$ & 52.74    & \cellcolor{mycolor}62.73$^{+9.99}$  & 54.68    & \cellcolor{mycolor}62.01$^{+7.33}$ \\
4$w$4$a$ & 63.39    & \cellcolor{mycolor}66.97$^{+3.58}$  & 65.80    & \cellcolor{mycolor}68.08$^{+2.28}$ & 66.81    & \cellcolor{mycolor}68.79$^{+1.98}$  & 66.94    & \cellcolor{mycolor}68.77$^{+1.83}$ \\
5$w$5$a$ & 66.12    & \cellcolor{mycolor}68.96$^{+2.84}$  & 69.26    & \cellcolor{mycolor}70.14$^{+0.88}$ & 69.93    & \cellcolor{mycolor}70.05$^{+0.12}$  & 69.75    & \cellcolor{mycolor}70.21$^{+0.46}$ \\ \hline 
\end{tabular}%
}
\caption{Accuracy of existing SOTA methods with the integration of \textit{GranQ}. GDFQ and AdaDFQ focus on data generation, whereas Qimera+AIT and AdaDFQ+AKT primarily enhance quantized model training.}
\label{tab4}
\vspace{-0.8em}
\end{table*}

\begin{figure*}[t]
\centering
\begin{minipage}[t]{0.44\textwidth}
    \centering
    \includegraphics[width=\textwidth]{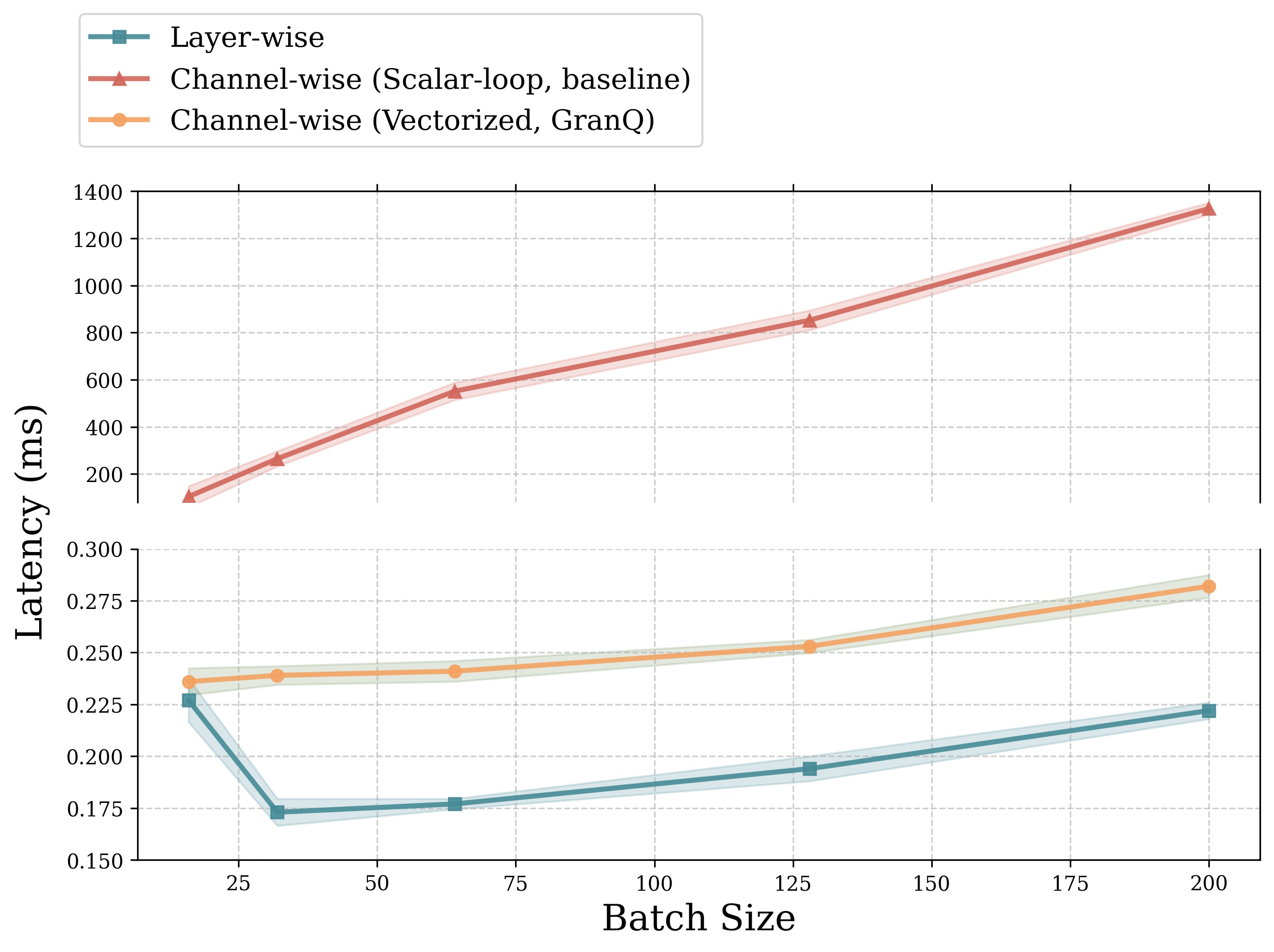}
    \caption{Latency of ResNet-20 quantization across batch sizes on CIFAR-100 with 3-bit setting.}
    \label{fig4}
\end{minipage}
\hfill
\begin{minipage}[t]{0.44\textwidth}
    \centering
    \includegraphics[height=5cm, width=\textwidth]{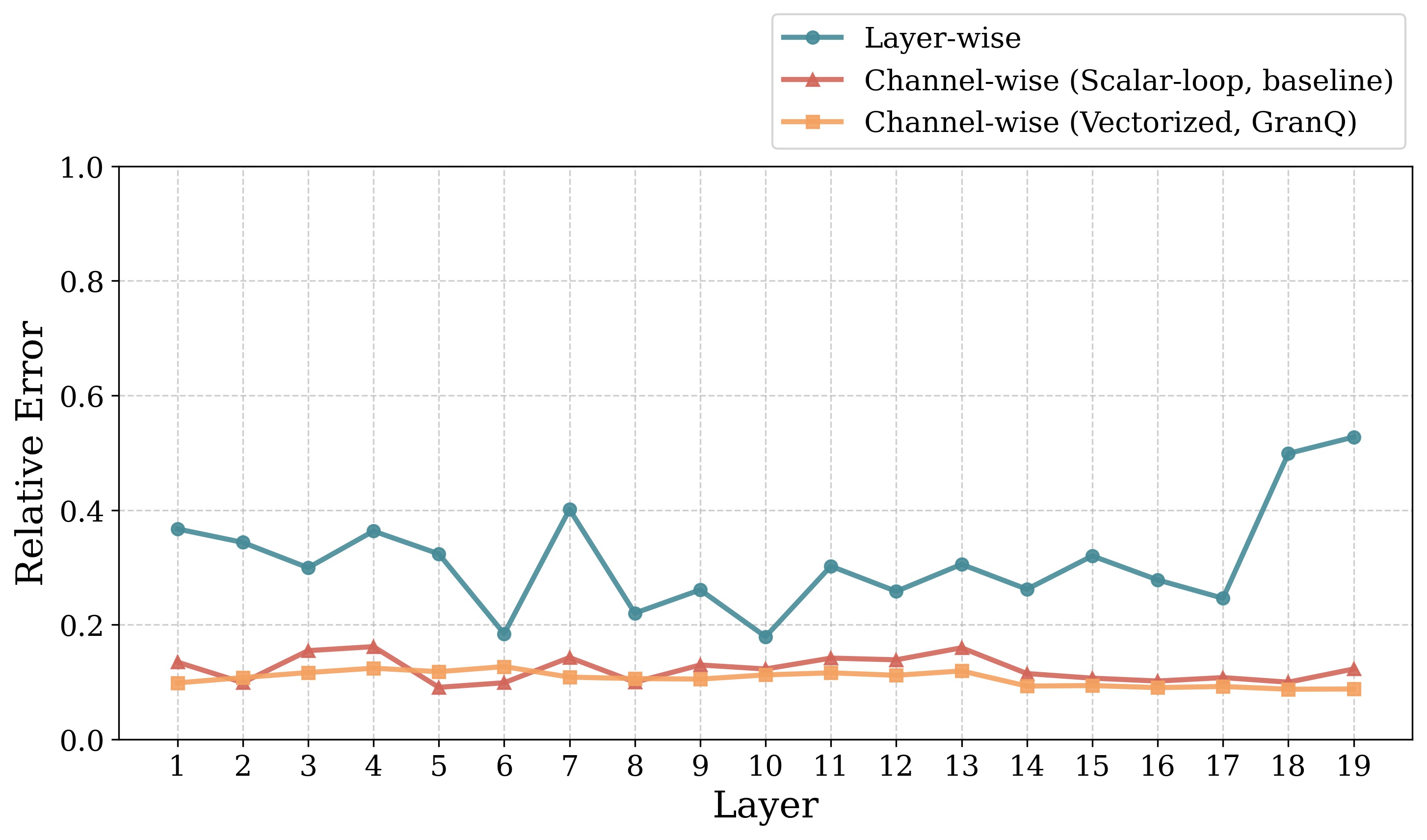}
    \caption{Relative quantization error across layers in ResNet-20 with 3-bit quantization on CIFAR-100.}
    \label{fig5}
\end{minipage}

\end{figure*}

\subsection{Performance Evaluation}
We evaluated the performance of \textit{GranQ} against SOTA ZSQ methods, with results summarized in Table \ref{tab3}. All comparison experiments were conducted under 3, 4, and 5-bit quantization settings.

\textbf{CIFAR-10/100.} \textit{GranQ} consistently achieved the highest accuracy across all bitwidths. For CIFAR-10, it attained accuracies of 94.06\% (5-bit), 93.52\% (4-bit), and 91.37\% (3-bit). For CIFAR-100, the results were 70.05\% (5-bit), 68.79\% (4-bit), and 62.73\% (3-bit). Notably, \textit{GranQ} outperformed SynQ \cite{30}, the previous SOTA, by +5.45\% in the CIFAR-100 3-bit setting. This result demonstrates its ability to effectively overcome the limitations of conventional low-bitwidth quantization techniques.
Overall, \textit{GranQ} consistently outperformed existing methods across all bitwidths in both CIFAR-10 and CIFAR-100, with particularly strong improvements in the 3-bit quantization setting. Remarkably, in the CIFAR-10 5-bit setting, \textit{GranQ} even surpassed the performance of the FP model. These results suggest that we can effectively apply \textit{GranQ} to small- and medium-scale datasets.

\textbf{ImageNet.} In the ImageNet experiments, \textit{GranQ} achieved competitive performance across various bitwidths. Specifically, for ResNet-18, it attained top accuracies of 70.39\% (4-bit) and 71.31\% (5-bit). For ResNet-50, it achieved the highest accuracies of 76.63\% (4-bit) and 71.31\% (5-bit). Additionally, in the MobileNetV2 setting, \textit{GranQ} achieved SOTA performance across all bitwidths (3, 4, and 5-bit).
In the 3-bit setting, \textit{GranQ} achieved the second-best accuracies, with 64.41\% on ResNet-18 and 70.76\% on ResNet-50. These results are -3.77\% and -3.23\% lower than those of GenQ, respectively. GenQ \cite{46} employs a pre-trained diffusion model for data generation, resulting in slower speeds than GAN-based methods. In contrast, \textit{GranQ} focuses on enhancing the quantization mechanism itself. Although GenQ's code is not publicly available, precluding its inclusion in our experiments, our framework is compatible and could potentially benefit from its data generation approach if it becomes accessible.



\begin{table*}[t]
\centering
\resizebox{0.8\textwidth}{!}{

\begin{tabular}{lccccc}
\hline
\textbf{Method} & 
\textbf{\begin{tabular}[c]{@{}c@{}}Quantization\\(Vectorized)\end{tabular}} & 
\textbf{\begin{tabular}[c]{@{}c@{}}Pre-Scaling\\(Parallel Accumulation) \end{tabular}} & 
\textbf{Accuracy (\%)} & 
\textbf{\begin{tabular}[c]{@{}c@{}}Quantization\\Latency (ms)\end{tabular}} 
 & 
\textbf{\begin{tabular}[c]{@{}c@{}}Training\\Time (sec/epoch)\end{tabular}} \\
\hline
Layer-wise & \checkmark & & 49.98 & 0.227 & 10.92 \\
Channel-wise (Baseline) & \checkmark & & 62.68  & 103.671 & 1737.54 \\
\rowcolor{mycolor} Channel-wise (\textbf{\textit{GranQ}}) & \checkmark & \checkmark & \textbf{62.73} & \textbf{0.236} & \textbf{12.43}\\
\hline
\end{tabular}
}
\caption{Ablations on the impact of pre-scaling in quantization. Quantization latency is measured with a batch size of 16 using AdaDFQ on ResNet-20 (CIFAR-100, 3-bit), representing the total time for data conversions required by the quantization scheme.}
\label{tab5}
\vspace{-1em}
\end{table*}

\subsection{Ablation Study}

\subsubsection{Effectiveness Evaluation}

\textit{GranQ} consistently demonstrates performance improvements when applied to various ZSQ methods. As summarized in Table \ref{tab4}, \textit{GranQ} achieves steady performance improvements when integrated with existing SOTA ZSQ techniques.

First, we analyzed the impact of \textit{GranQ} on data synthesis-based quantization methods, specifically GDFQ \cite{16} and AdaDFQ \cite{22}. The integration of \textit{GranQ} into GDFQ \cite{16} led to a notable improvement, with accuracy rising from 47.61\% to 59.04\% (+11.43\%) in the 3-bit setting and from 63.39\% to 66.97\% (+3.58\%) in the 4-bit setting. Similarly, AdaDFQ \cite{22} exhibited improvements of +9.99\% in the 3-bit setting (from 52.74\% to 62.73\%) and +1.98\% in the 4-bit setting (from 66.81\% to 68.79\%). These results suggest that \textit{GranQ} effectively reduces quantization errors when combined with data synthesis-based quantization methods, leading to enhanced model performance.

Furthermore, \textit{GranQ} also exhibits significant improvements in methods focused on training quantized models, such as Qimera+AIT \cite{27} and AdaDFQ+AKT \cite{29}. For Qimera+AIT \cite{27}, the 3-bit accuracy increased from 45.70\% to 60.42\% (+14.72\%). Similarly, for AdaDFQ+AKT \cite{29}, the accuracy improved from 54.68\% to 62.01\% (+7.33\%). These findings demonstrate that \textit{GranQ} is not only effective in data synthesis-based methods but also enhances performance during the model training process.





\subsubsection{Efficiency Evaluation}

While fine-grained activation quantization often incurs high computational cost due to per-channel scaling, \textit{GranQ} maintains efficiency by vectorizing the scaling computation. By reshaping activation tensors, our method eliminates the overhead of iterative channel-wise operations. To assess this in practice, we measured quantization latency across various batch sizes (16, 32, 64, 128, and 200). Figure~\ref{fig4} compares three methods: layer-wise quantization, channel-wise quantization with scalar-loop scaling, and our method (\textit{GranQ}), which efficiently parallelizes scaling across channels.
The two channel-wise methods perform per-channel quantization. The conventional method computes scaling factors sequentially, whereas \textit{GranQ} pre-applies them in a vectorized manner across channels, enabling pre-scaling for efficient parallel accumulation. A key observation from Figure~\ref{fig4} is that \textit{GranQ} achieves substantial accuracy improvement with only a minimal latency overhead compared to conventional layer-wise quantization. Furthermore, Table~\ref{tab2} and Figure~\ref{fig5} show that \textit{GranQ} substantially reduces quantization error and more effectively preserves activation information. Additionally, \textit{GranQ} demonstrates that per-channel quantization can be efficiently executed by vectorizing scaling factor computation. As shown in Table~\ref{tab5}, our method achieves a quantization latency of 0.236 ms, which is comparable to the layer-wise baseline (0.227 ms) while significantly outperforming the conventional channel-wise method (103.671 ms) that suffers from scalar-loop overhead.

\section{Discussion}
\label{sec:discussion}

\subsection{Why \textit{GranQ} is Effective?}

\begin{figure}[t]
\centering
\begin{subfigure}[t]{0.36\textwidth}
    \centering
    \includegraphics[width=\textwidth]{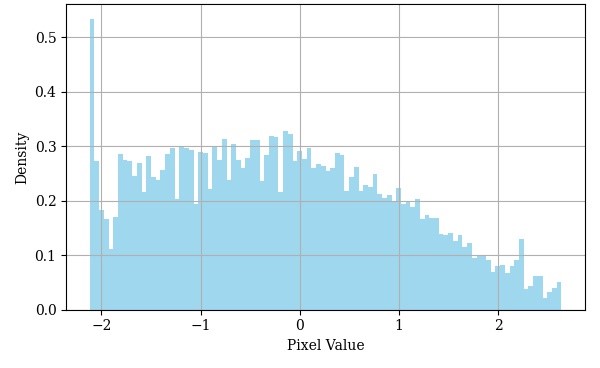}
    \caption{Original inputs (Skewness: 0.2599)}
    \label{fig:real}
    \vspace{0.6em}
\end{subfigure}
\hfill
\begin{subfigure}[t]{0.36\textwidth}
    \centering
    \includegraphics[width=\textwidth]{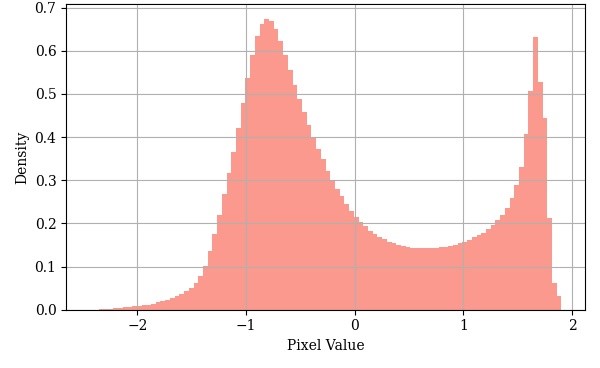}
    \caption{Synthetic inputs (Skewness: 0.4877)}
    \label{fig:synthetic}
\end{subfigure}

\caption{Pixel distribution comparison between (a) original and (b) synthetic inputs on ImageNet. Synthetic inputs generated by AdaDFQ~\cite{22} exhibit a rightward shift with higher skewness.}

\label{fig6}
\vspace{-1.9em}
\end{figure}

In layer-wise quantization, representing the entire activation range with a single scaling factor makes it difficult to reflect fine-grained distribution changes. These limitations become particularly severe in low-bit settings, especially when synthetic data exhibit distributional shifts, a common scenario in ZSQ. Figure~\ref{fig6} highlights the severity of this issue. The synthetic data central to ZSQ exhibits significantly higher skewness compared to the original data, resulting in a greater prevalence of extreme values, or outliers. This characteristic poses a significant challenge for conventional methods. With a single, global scaling factor, a few extreme outliers force the quantization range to become excessively wide, which in turn compresses the majority of the distribution into a limited number of discrete levels and leads to severe quantization error.

To overcome this fundamental challenge, \textit{GranQ} employs a granular, adaptive scaling mechanism. It pre-computes per-channel scaling in a vectorized form, enabling precise representation of skewed distributions while allowing for efficient parallel accumulation. By integrating this scaling into the quantization phase, we remove runtime scaling overhead, thus achieving both the precision required for low-bit ZSQ and the efficiency of parallel execution.

\subsection{Limitation and Future-work}

\textit{GranQ} efficiently addresses channel-wise activation range in zero-shot QAT. However, PTQ does not apply quantization during training and thus differs fundamentally from QAT. As future work, we will extend \textit{GranQ} to granular quantization in PTQ settings.



\section{Conclusion}
\label{sec:conclusion}

ZSQ faces a critical trade-off between accuracy and efficiency. The reliance on synthetic data often introduces distributional shifts, which demand fine-grained, per-channel scaling for activations. However, this approach incurs significant computational overhead, limiting its practical application. To resolve this, we propose \textit{GranQ}, a novel quantization scheme that pre-computes and vectorizes per-channel scaling factors. This allows for both precise, fine-grained quantization and highly efficient parallel accumulation. 

Our extensive experiments demonstrate that  \textit{GranQ} resolves the critical accuracy-latency trade-off, achieving SOTA accuracy while maintaining the low latency of layer-wise methods. Ultimately, our work provides a practical solution to the core activation challenge in ZSQ, enabling the deployment of highly accurate, low-bit models without original data.

{
    \small
    \bibliographystyle{ieeenat_fullname}

}

 \end{document}